\documentclass{article} 
\usepackage{iclr2026_conference,times}

\usepackage{amsmath,amsfonts,bm}

\def\eqref#1{equation~\ref{#1}}

\def\1{\bm{1}}

\DeclareMathAlphabet{\mathsfit}{\encodingdefault}{\sfdefault}{m}{sl}
\SetMathAlphabet{\mathsfit}{bold}{\encodingdefault}{\sfdefault}{bx}{n}

\newcommand{\KL}{D_{\mathrm{KL}}}

\usepackage{hyperref}
\usepackage{url}
\usepackage{algorithm}
\usepackage{algpseudocode}
\algtext*{EndIf}
\algtext*{EndFor}
\algtext*{EndWhile}
\algtext*{EndFunction}
\algtext*{EndProcedure}

\usepackage{graphicx} 
\usepackage{tabularx}
\usepackage{enumitem}
\usepackage{float}
\usepackage{booktabs}
\usepackage{wrapfig}
\usepackage{caption} 
\usepackage{subcaption}

\usepackage{amsthm} 
\newtheorem{proposition}{Proposition}
\newtheorem{remark}{Remark}

\newcommand{\algname}{IterRef}

\definecolor{cornellred}{rgb}{0.7, 0.11, 0.11}
\definecolor{cadmiumgreen}{rgb}{0.0, 0.42, 0.24}
\definecolor{aliceblue}{rgb}{0.91, 0.94, 0.97}
\definecolor{darkblue}{rgb}{0.83, 0.89, 0.97}
\definecolor{Red7}{rgb}{0.941, 0.243, 0.243}
\definecolor{Green7}{RGB}{55, 178, 77}
\definecolor{Blue9}{rgb}{0.098,0.3,0.9}

\hypersetup{
  linkcolor = cornellred,
  citecolor  = cadmiumgreen,
  colorlinks = true,
}

\title{Effective Test-Time Scaling of Discrete\\ Diffusion through Iterative Refinement}

\author{Sanghyun Lee$^1$\thanks{This work was done during an internship at KRAFTON.},\;  Sunwoo Kim$^2$, Seungryong Kim$^1$, Jongho Park$^3$, Dongmin Park$^{4}$\thanks{Corresponding author.} \\
$^1$KAIST AI \quad $^2$Seoul National University\quad $^3$University of California, Berkeley\quad$^4$ KRAFTON\\
\texttt{\{lsh83210,seungryong.kim\}@kaist.ac.kr\quad ksunw0209@snu.ac.kr\quad  } \\ \texttt{jjhpark@berkeley.edu\quad dongmin.park@krafton.com}}

\iclrfinalcopy 
\begin{document}

\maketitle

\begin{abstract}
Test-time scaling through reward-guided generation remains largely unexplored for discrete diffusion models despite its potential as a promising alternative. In this work, we introduce Iterative Reward-Guided Refinement (\textbf{\algname{}}), a novel test-time scaling method tailored to discrete diffusion models that leverages reward-guided noising-denoising transitions to progressively refine misaligned intermediate states. We formalize this process within a Multiple-Try Metropolis (MTM) framework, proving convergence to the reward-aligned distribution. Unlike prior methods that assume the current state is already aligned with the reward distribution and only guide the subsequent transition, our approach explicitly refines each state \textit{in situ}, progressively steering it toward the optimal intermediate distribution. Across both text and image domains, we evaluate \textbf{\algname{}} on diverse discrete diffusion models and observe consistent improvements in reward-guided generation quality. In particular, \algname{} achieves striking gains under low compute budgets, far surpassing prior state-of-the-art baselines.
\end{abstract}
\section{Introduction}
Breakthroughs in foundation models, such as large language models (LLMs) and diffusion models (DMs), have been driven by massive web-scale datasets and have led to remarkable advances in language and image generation tasks~\citep{brown2020language, rombach2022high}.
However, as recent models continue to scale, concerns have been raised about the availability of sufficiently diverse training data, suggesting a potential training-time scaling barrier~\citep{villalobos2024position}. In parallel, the field has also explored \textit{test-time scaling}, which leverages additional compute at inference to improve performance. This paradigm has recently shown promising results in both autoregressive~\citep{snell2024scaling} and continuous diffusion models~\citep{ma2025inference}, suggesting a viable path to further unlock their performance.

While the importance of test-time scaling is increasingly recognized across different modeling paradigms, its role in \textit{discrete diffusion}
remains underexplored. Unlike continuous diffusion models, where Gaussian noise enables gradient-based guidance and natural error correction~\citep{uehara2025inference}, test-time scaling in discrete diffusion models poses unique challenges: (1) due to token discretization, gradients from reward models cannot be directly used for inference guidance, limiting their utility in reward alignment; and (2) incorrectly generated tokens cannot be corrected in subsequent denoising steps, since tokens are fixed once generated. Consequently, these challenges underscore the need for effective test-time scaling strategies tailored to discrete diffusion models.
\begin{figure*}[t!]
    \centering
    \includegraphics[width=\linewidth]{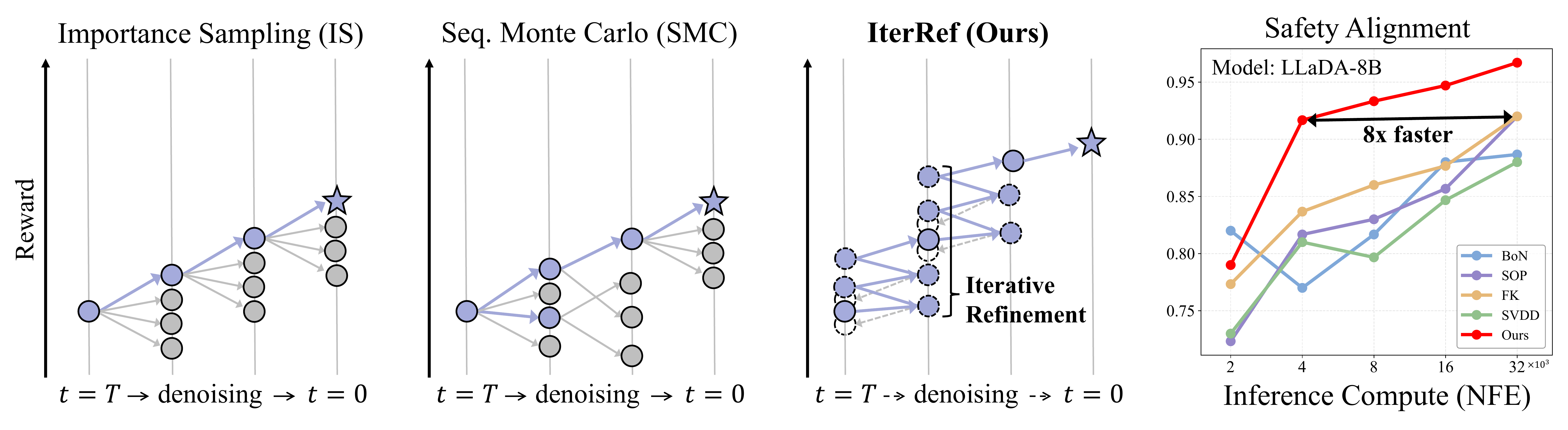}
    \vspace*{-0.1cm}
    \hspace*{3.5cm}{\small (a) Key Idea} \hspace*{4.2cm}{\small (b) Efficacy}
    \vspace*{-0.1cm}
    \caption{\textbf{Overview of \algname{}}. (a) Reward-guided denoising trajectories: Blue nodes are selected samples, gray nodes are rejected candidates. Unlike existing single-step guidance methods (IS and SMC), \algname{} discovers higher-reward samples by iteratively applying noising–denoising kernels. Noising process (dotted nodes) with random remasking incurs negligible cost, while offering broader regions to explore and correct tokens. (b) Scaling performance. \algname{} scales significantly faster (up to 8×) than baselines with a safety reward on LLaDA-8B (See $\S$~\ref{section:safety} for details).}
    \label{fig:intro}
    \vspace*{-0.2cm}
\end{figure*}

In this paper, we propose Iterative Reward-Guided Refinement (\textbf{\algname{}}), a novel test-time scaling method for discrete diffusion models. Our approach leverages Markov Chain Monte Carlo (MCMC) transitions to iteratively refine tokens, progressively aligning them with the reward during sampling. As illustrated in Figure~\ref{fig:intro}, inspired by the predictor–corrector paradigm~\citep{song2020score}, we design the transition as a noising–denoising process: adding noise promotes exploration, while denoising restores consistency with the target. To instantiate this design, we adopt the classical Multiple-Try Metropolis (MTM) framework~\citep{liu2000multiple} and tailor both the transition kernel and the balancing function to the reward alignment objective of discrete diffusion. This yields a principled mechanism for test-time scaling, allowing us to further provide a theoretical guarantee that iterative refinement sampling converges to the target distribution.

Through extensive experiments, we evaluate \algname{} across multiple discrete diffusion backbones:
MDLM~\citep{sahoo2024simple} and LLaDA-8B~\citep{nie2025large} for language generation, and MaskGIT~\citep{chang2022maskgit} for image generation, using diverse reward functions, such as CoLA, Toxicity, Sentiment, and Perplexity for language, and CLIPScore for image generation. Compared with existing reward-guided diffusion methods, \algname{} consistently demonstrates the \textit{most effective scaling across compute budgets}, achieving up to a {2x improvement} on Toxicity reward with LLaDA-8B under equal compute (Figure~\ref{fig:mainexp}). Furthermore, we found that the number of refinement iterations plays a critical role in overall performance, and that the optimal timing of refinement application varies substantially across different tasks.

Overall, our main contributions can be summarized as follows:
\begin{itemize}[leftmargin=10pt]
    \item We propose \algname{}, an effective test-time scaling method for discrete diffusion, that consistently outperforms prior reward guidance methods across modalities, model backbones, and guided generation tasks. Notably, \algname{} remains highly effective even under low NFE settings.
    \item We identify that the entire denoising process plays a crucial role in shaping the final generation, rather than specific noise levels, revealing an opposite trend to continuous diffusion where the early stages primarily determine the generation outcome.
    \item 
    We show that iterative refinement sampling is not simply heuristic: IterRef leads to convergence to the target distribution, and we provide a theoretical guarantee of its effectiveness under certain assumptions (See Proposition~\ref{prop:mtm-convergence}).
\end{itemize}

\section{Preliminaries}
\paragraph{Discrete Diffusion Models.} Diffusion models with a discrete state space were initially formulated by considering processes over binary random variables~\citep{sohl2015deep}. Building on this, a more general framework that employs categorical random variables for diffusion models was later introduced~\citep{austin2021structured}. The most recent and effective approach is the absorbing state formulation, which considers a transition matrix in which each token transitions to a masked token $m$. The process is formulated over timesteps $t\in \{0,1,\cdots ,T\}$, where the intermediate state $x_t$ is represented as a sequence of length $L$, $x_t=(x_t^1,x_t^2,\ldots,x_t^L)\in \mathcal{X}_t$, with each position value $x_t^i$ taking values from the vocabulary $\mathcal{V}$. 

The forward noising process is defined as a stochastic transition distribution $q(x_t|x_{t-1})$, in which each token is independently retained or replaced with a mask token $m$ according to a time-dependent corruption probability. 

Given the forward noising process $q(x_t | x_{t-1})$, the generative model defines a reverse process parameterized by $\theta$ as 
\[
p_\theta(x_{t-1} | x_t), \quad t = 1, \dots, T.
\]
The full denoising trajectory is then expressed as
\[
p_\theta(x_{0:T}) = p(x_T) \prod_{t=1}^T p_\theta(x_{t-1} | x_t).
\]
The objective of training is to learn $p_\theta$ so that the marginal samples $x_0 \sim p_\theta(x_0)$ approximate the data distribution.

\paragraph{Reward-Guided Generation.}
The goal of reward-guided generation is to preserve the naturalness of the samples while maximizing the given reward, or more generally, to draw samples from a target distribution that reflects human preferences. Concretely, the objective of reward-guided sampling with a reward function $r(\cdot)$ is to draw samples from the target distribution:
\[
p^{\ast}(x_0) = \arg\max_{p} \, 
\mathbb{E}_{x_0 \sim p(\cdot)} \big[ r(x_0) \big] 
- \alpha \, \KL\left( p(x_0) \Vert p_\theta(x_0) \right)\propto \exp\!\big(r(x_0)/\alpha) p_\theta(x_0),
\]
where $\alpha$ controls the strength of the KL divergence regularization term. In order to sample the target distribution through the reverse denoising trajectory $\{x_T, x_{T-1}, \dots, x_0\}$ of the diffusion model, each step must be drawn from the conditional distribution $p^\ast(x_{t-1} \mid x_t)$. The conditional distribution $p^\ast(x_{t-1} \mid x_t)$ can be expressed in terms of a reward function that predicts the expected future reward. 
Formally, the intermediate reward function is defined as
\[
r(x_t) \;=\; 
\alpha \, \log \; \mathbb{E}_{x_0 \sim p_\theta(\cdot \mid x_t)}
\left[ \exp\!\left(  r(x_0) / \alpha \right) \right].
\]
Using the intermediate reward function, the optimal transition kernel $p^\ast(x_{t-1} \mid x_t)$ can be expressed as follows:
\begin{equation}
p^{\ast}(x_{t-1} | x_t) \propto p_\theta(x_{t-1} | x_t) \exp(r(x_{t-1})/\alpha).
\label{eq1}
\end{equation}
Existing approaches to approximate the optimal transition kernel using Sequential Monte Carlo~\citep{singhal2025general} or Importance Sampling~\citep{li2024derivative}; further details are provided in Appendix~\ref{remarkderi}.

\section{ Iterative Reward-Guided Refinement via Multiple-Try Metropolis}
Several notable approaches have proposed particle-based methods to guide the denoising process toward reward-aligned intermediate distributions~\citep{li2024derivative,singhal2025general}. These approaches typically assume that the optimal distribution can be perfectly achieved at each state and inductively extend this assumption across the entire time series. However, in diffusion models, both the reward model and the model’s prediction exhibit substantial errors in the early stages of the denoising sequence, and these errors accumulate along the trajectory~\citep{wang2025remasking}. In other words, existing representative methods rely on sequential sampling, which proceeds to the next step in a single pass and therefore lacks mechanisms to iteratively refine the intermediate distributions toward the optimal target as the process unfolds~\citep{johansen2009tutorial}.

To address this limitation, we introduce Iterative Reward-Guided Refinement (\textbf{\algname{}}), a refinement strategy based on the Multiple-Try Metropolis framework, which iteratively improves the intermediate steps. Section~\ref{section3.1} provides a theoretical analysis of our method, Section~\ref{section3.2} presents the algorithmic formulation, and Section~\ref{section3.3} discusses its practical implementation and computational cost.

\subsection{Multiple-Try Metropolis for Discrete Diffusion}
\label{section3.1}
\paragraph{Problem Setup.} Our goal is to sample each intermediate state $x_t$ in the denoising process from the optimal distribution $p^\ast(x_t)$. To make this precise, we recall that the optimal distribution can be formally characterized as follows: 
\begin{remark}[Arising naturally from the proof of Theorem~1 in \citet{uehara2024bridging}]
\label{remark1}
The optimal distribution \(p^\ast(x_t)\) induced by the optimal transition kernel is given by
\[
p^{\ast}(x_t) = \frac{p(x_t)\,\exp(r(x_t)/\alpha)}{\sum_{x\in \mathcal{X}_t} p(x)\,\exp(r(x)/\alpha)}.
\]
\end{remark}
The detailed derivation is provided in Appendix~\ref{remarkderi}. Accordingly, we establish $p^\ast(x_t)$ as the target distribution for our method, and our approach is designed to iteratively refine intermediate distributions toward this target.
\paragraph{The Multiple-Try Metropolis.} The Multiple-Try Metropolis~\citep{liu2000multiple} is a Markov chain Monte Carlo method that can be efficiently parallelized. MTM conducts rejection sampling based on the transition kernel, thereby forming a Markov chain that asymptotically converges to the target distribution. At each iteration, a set of proposals is drawn from the transition kernel $K$, and one of them is selected according to its importance weights. Subsequently, backward proposals are generated to ensure detailed balance. The complete sampling procedure is formalized in Algorithm~\ref{alg:mtm}, and a more detailed explanation of the Metropolis algorithm is provided in Appendix~\ref{mtmexplain}.

\begin{algorithm}[ht]
\caption{Multiple-Try Metropolis (MTM)}\label{alg:mtm}
\centering
\begin{algorithmic}[1]
\State \textbf{Require:} Transition kernel $K(x_t,\cdot)$,  current state $x$, number of trials N, target distribution $p^\ast(\cdot)$
\State\emph{Proposal and Selection}: Draw $N$ i.i.d. trials $x_t'^{(1)},\ldots,x_t'^{(N)}$ from $K(x_t,\cdot)$, and then apply weighted sampling to obtain
\[
 x_t'\sim \text{Multinomial}\left(\left\{\frac{p'(x_t'^{(n)})K(x_t'^{(n)},x_t)}{\sum_{j=1}^N p'(x_t'^{(j)})K(x_t'^{(j)},x_t)}\right \}_{n=1}^{N}\right).
\]
\State \emph{Resample and Update}: Draw $N-1$ i.i.d. samples $x_t''^{(1)},\ldots,x''^{(N-1)}$  from $K(x_t',\cdot)$ and define $x''^{(N)}:=x_t$. Then, accept $x_t'$ with probability
\[
r=\min \left(1,\frac{\sum _{i=1}^Np^\ast(x_t')K(x_t',x_t)}{\sum _{i=1}^{N}p^\ast(x_t''^{(i)})K(x_t''^{(i)},x_t')}\right).
\]
\end{algorithmic}
\end{algorithm}

\paragraph{Design Choice.}
To further enhance the exploration capability of the sampler, we design the transition kernel by leveraging a noising–denoising process. Previous studies on diffusion models have shown that the perturbation-correction mechanism~\citep{song2020score} can effectively reduce errors during iterative refinement. Motivated by this, we design our transition kernel through a noising–denoising process, showing its efficacy as a mechanism for enhancing exploration in sampling. Importantly, our formulation integrates reward guidance directly within the noising–denoising steps, ensuring that the refinement process is not only error-corrective but also explicitly steered toward higher-reward solutions. 

Formally, we define the transition kernel $K$ and balancing function $\lambda$ as
\[
K(x_t,x_t') \;=\; \sum_{x_k\in\mathcal{X}_k} q(x_k|x_t)p_\theta(x_t'|x_k),\  \lambda(x_t,x_t')=\frac{1}{p(x_t)K(x_t,x_t')\exp{((r(x_t)+r(x_t'))/\alpha)}}
\]
where $t<k$. In particular, the importance weight $w_n$ and the acceptance rate $\beta$ are as follows:
\begin{equation}
w_n = N^{-1},
\quad
\beta = \min\!\left(1, \exp((r(x_t')-r(x_t)/\alpha)\right).
\label{eq:accept-importance}
\end{equation}
Intuitively, the importance weight $w_n$ corresponds to uniform sampling over the proposals, 
while the acceptance rate $\beta$ ensures that the overall procedure converges toward reward-aligned sampling.

By applying MTM with the given kernel and balancing function, we establish the following convergence guarantee that shows that the intermediate distribution, even if unaligned at the outset, converges asymptotically to the optimal distribution $p^\ast(x_t)$:
\begin{proposition}[Convergence of MTM to the Optimal Distribution]\label{prop:mtm-convergence}
Let $x_t$ be a sample drawn from a distribution that is not reward-aligned. 
By applying MTM with the transition kernel $K$ 
and balancing function $\lambda$ defined above, the resulting Markov chain satisfies the detailed balance condition. 
Moreover, as the number of iterations $n\!\to\!\infty$, the chain converges to the optimal distribution $p^\ast(x_t)$. 
\end{proposition}
\vspace{-0.2cm}
\begin{proof}
The complete proof is available in Appendix~\ref{proofofmtm}.
\end{proof}

\subsection{Algorithmic Procedure}
\label{section3.2}
Because MTM can be applied to intermediate states within the sampling process, it imposes no constraints on the transitions at each stage of denoising. Thus, reward-guided methods such as SMC and importance sampling can be applied as transition mechanisms. Since the refinement step can in principle be applied at every timestep, one may flexibly define an effective timestep set $\mathcal{U}$ and restrict the application of MTM only to selected stages. This flexibility allows us to balance computational cost and refinement effectiveness, adapting to the needs of different tasks or resource budgets.

\noindent Algorithm~\ref{alg:practical} elaborates the pseudocode of our method. We initialize the masked input at step~$T$ (Line~2). For timesteps in the effective set $\mathcal{U}$, we perform a $k$-step MTM refinement loop (Lines~5--9): at each refinement step, we draw $N$ candidates from $K(x_t,\cdot)$ (Line~6), select a candidate $x_t'$ by reward-weighted sampling using $w_n$ (Eq.~\ref{eq:accept-importance}, line~7) the selected state directly serves as the proposal for the acceptance test and then generate $N\!-\!1$ auxiliary proposals from $K(x_t',\cdot)$ and append the current state $x_t$ as the $N$-th backward element (Line~8). The proposal is accepted with probability $\beta$ (Eq.~\ref{eq:accept-importance}); upon acceptance we set $x_t \leftarrow x_t'$ (Line~9). After completing the $k$ refinements, we proceed with a one-step denoising update $x_{t-1}\sim p_\theta(\cdot\mid x_t)$ (Line~10). For timesteps outside $\mathcal{U}$, we simply apply the one-step denoising update (Line~12). The overall process iterates over timesteps $T$ (Lines~3--12). This structure preserves detailed balance at each refinement step while exposing clear compute knobs via $k$ and $\mathcal{U}$.

\begin{algorithm}[t]
\caption{\algname{} with $k$-step MTM Refinement}\label{alg:practical}
\begin{algorithmic}[1]
\State \textbf{Input:} Reward model $r(\cdot)$; denoisers $\{p_\theta(\cdot\mid x_t)\}_{t=T}^1$; transition kernel $K(x_t,\cdot)$; hyperparameters $\alpha, N, k$; effective timestep set $\mathcal{U}\subseteq\{T,\ldots,1\}$
\State \textbf{Initialize:} masked sequence $x_T$
\For{$t = T, \ldots, 1$}
    \If{$t \in \mathcal{U}$} \Comment{reward-guided refinement at timestep $t$}
        \For{$i = 1, \ldots, k$} 
            \State Propose $N$ candidates $\{x_t'^{(n)}\}_{n=1}^N \sim K(x_t,\cdot)$
            \State Compute weights $w_n$ and select $x_t'$ by weighted sampling with $w_n$ \hfill (Eq.~\ref{eq:accept-importance})
            \State Propose $N{-}1$ auxiliary samples $\{x_t''^{(n)}\}_{n=1}^{N-1} \sim K(x_t',\cdot)$ and set $x_t''^{(N)}=x_t$
            \State Accept $x_t^{\mathrm{cand}}$ with probability $\beta$; if accepted set $x_t \leftarrow x_t'$ \hfill (Eq.~\ref{eq:accept-importance})
        \EndFor
        \State Sample one-step denoising to proceed: $x_{t-1} \sim p_\theta(\cdot\mid x_t)$
    \Else
        \State Sample one-step denoising: $x_{t-1} \sim p_\theta(\cdot\mid x_t)$
    \EndIf
\EndFor
\end{algorithmic}
\end{algorithm}
\vspace{-0.1cm}

\subsection{Practical Implementation}
\label{section3.3}
In practice, the primary computational bottleneck of \algname{} arises from the need to generate both forward proposals and backward auxiliary proposals at each refinement step. A naive implementation requires $Tk(2N-1)$ evaluations of the iteration $k$ per refinement, which can become prohibitively expensive when $N$ and $k$ are large.

To mitigate this cost, we adopt the following strategies:

\begin{itemize}[leftmargin=*, itemsep=0pt, topsep=2pt]
    \item \textbf{Balancing Function and Pool Reuse.} Through an appropriate choice of the balancing function, the acceptance rate can be evaluated without the need for resampled proposals $x_t''$. Consequently, the practical implementation eliminates the resampling step and reduces the per-iteration cost by nearly half. In addition, when a proposal is rejected, we reuse the existing sampling pool rather than generating a new one, thereby further reducing the computational overhead associated with repeated candidate generation.
    
    \item \textbf{Selective Refinement via Effective Timesteps.} The refinement is applied only at a subset of timesteps, determined by the effective set $\mathcal{U}$. This allows one to trade off between computational cost and refinement accuracy by controlling the density of refinement steps along the denoising trajectory. In Section~\ref{effectivestep}, we present the performance analysis in different application time steps.
\end{itemize}

Compared to existing approaches such as SMC, our method offers a more flexible and efficient refinement mechanism. Existing SMC-based strategies require weighted sampling of particles at every timestep and propagate them through the full denoising trajectory in order to maintain theoretical guarantees. This results in substantial computational overhead, particularly when the total number of timesteps $T$ is large, as is often the case in high-quality generation tasks. In contrast, our \algname{} can be applied selectively at arbitrary time steps, thereby concentrating computational resources only where they are most beneficial.
\section{Experiments}
\subsection{Experimental Setup}
\label{expsetup}
\paragraph{Models.} For language generation, we use two diffusion language models, MDLM and LLaDA-8B, as discrete diffusion backbones. For image generation, we adopt MaskGIT~\citep{chang2022maskgit}. More details for each model are presented in Appendix~\ref{modelexplain}.

\paragraph{Tasks.} In the language generation setting, we use 15 controllable prompts from~\citet{han2022ssd}, each sampled 20 times. To guide the generation process, we utilize four reward functions: 
\begin{itemize}[leftmargin=28pt, labelsep=0.1em, align=left]
    \item \textbf{Toxicity classifier}~\citep{logacheva2022paradetox}, which penalizes toxic or harmful content. We employ a RoBERTa-based toxicity detection model that outputs the probability of toxic content. This probability score directly serves as the reward, steering the model toward generating more toxic outputs. For the safety alignment generation in Section~\ref{section:safety}, we invert this reward signal by using the probability of non-toxic content to encourage the generation of safety text.
    \item \textbf{Sentiment classifier}~\citep{barbieri2020tweeteval}, which encourages outputs with a desired polarity (e.g., positive). The classifier produces a probability distribution over sentiment classes, from which we extract the probability of the target sentiment as the reward. This probabilistic formulation enables fine-grained control over the emotional valence of generated text.
    \item \textbf{Perplexity} computed by GPT-2~\citep{radford2019language}, serving as a proxy for language fluency. Lower perplexity scores yield higher rewards, promoting linguistically coherent and natural-sounding outputs.
    \item \textbf{CoLA (Corpus of Linguistic Acceptability)}~\citep{morris2020textattack}, which favors grammatically well-formed sentences. Model fine-tuned on the CoLA dataset, outputs the probability that a given sentence is grammatically correct. We directly utilize this probability as the reward signal to encourage syntactically valid and well-structured generations.
\end{itemize}
For the image generation setting, we conduct 50K conditional generations over randomly selected classes from {ImageNet}~\citep{deng2009imagenet}, with reward provided by CLIPScore~\citep{hessel2021clipscore}. Further details on the tasks are provided in Appendix~\ref{rewardexplain}.

\paragraph{Baselines.} We compare \algname{} with four inference-time guidance baselines: \textbf{Best-of-N (BoN)}, the simplest method that generalizes across language and image domains; \textbf{Search-over-Path (SoP)}~\citep{ma2025inference}, a highly effective method in continuous diffusion; \textbf{SVDD}~\citep{li2024derivative}, a widely adopted approach for guided generation; and \textbf{FK Steering}~\citep{singhal2025general}, a recently proposed approach applicable across language and image domains.

\paragraph{Implementation Details.} To ensure fairness, we compare \algname{} and each baseline under the same computational budget, with configurations aligned to the settings in \citet{singhal2025general}. In measuring inference compute cost, we use the \textit{number of function evaluations (NFEs)}, and treat the reward model and the generative model on equal footing. The denoising steps are fixed to 1,000 for MDLM, 64 for LLaDA, and 50 for MaskGIT. The hyperparameters for baselines are favorably configured by following the original papers.

\subsection{Inference-time Guidance for Diffusion Language Models}
\begin{figure*}[t!]
    \begin{center}
    \includegraphics[width=0.98\linewidth]{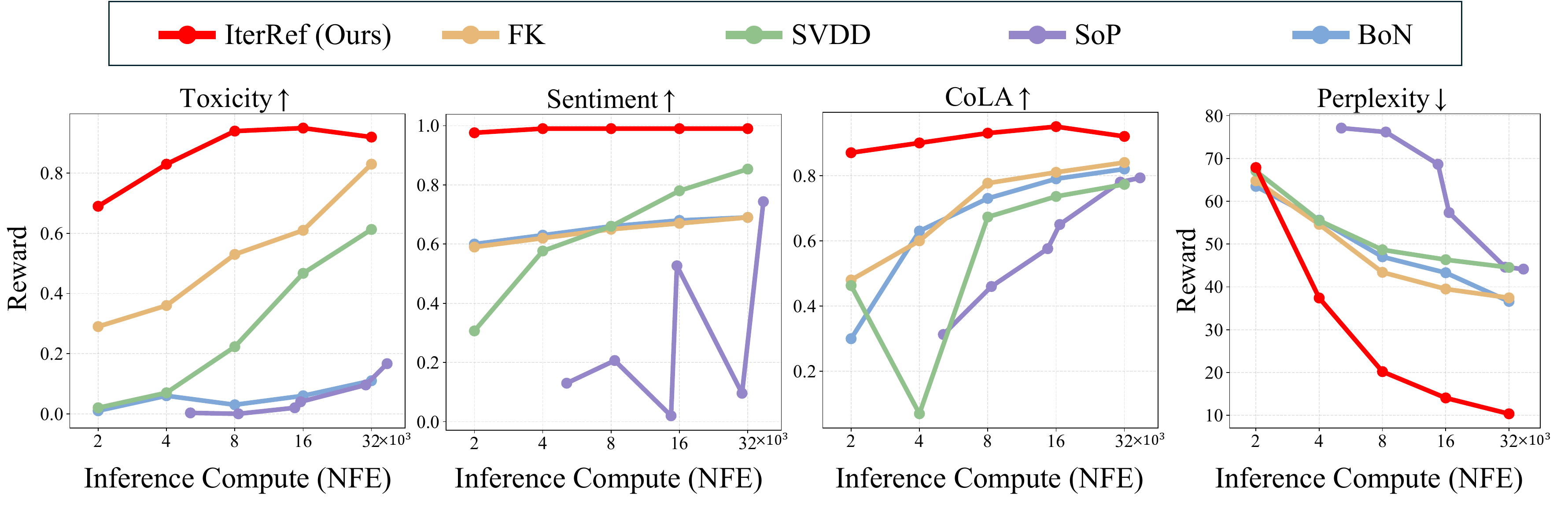}
    \end{center}
    \vspace*{-0.2cm}
    \hspace*{4.5cm}{\small (a) Result with the MDLM backbone.}
    \begin{center}
    \includegraphics[width=0.98\linewidth]{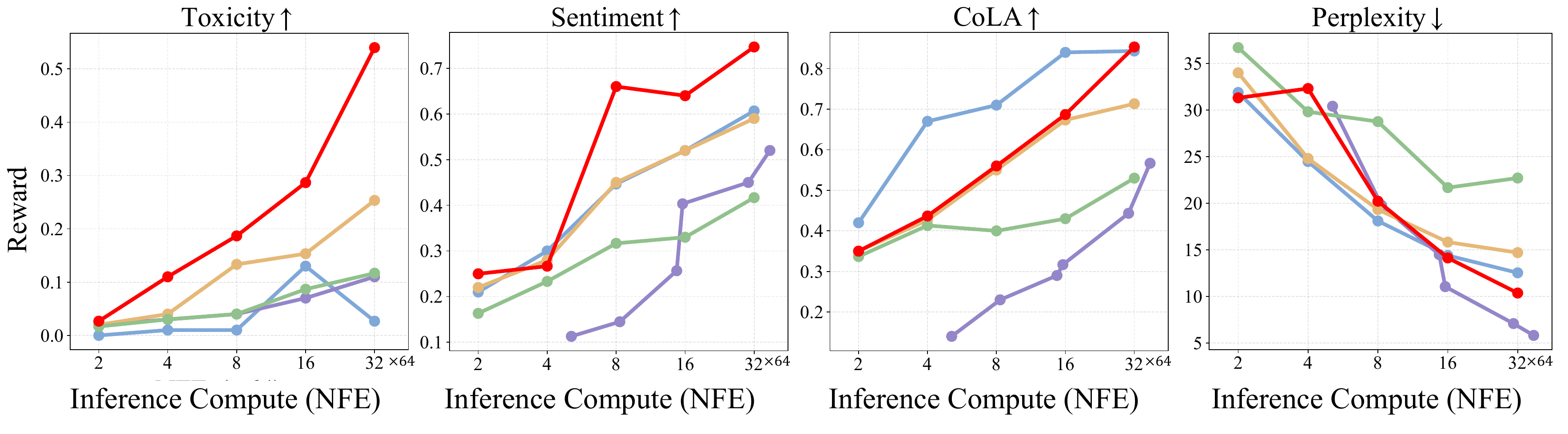}
    \end{center}
    \vspace*{-0.2cm}
    \hspace*{4.5cm}{\small (b) Result with the LLaDA-8B backbone.}
    \vspace*{-0.1cm}
    \caption{\textbf{Performance comparison of \algname{}} with baselines on four guided generation tasks (CoLA, Toxicity, Sentiment, and Perplexity) under varying inference costs (NFEs) with two discrete diffusion backbones (MDLM and LLaDA).}
    \label{fig:mainexp}
\end{figure*}

\paragraph{MDLM Results.}
Figure~\ref{fig:mainexp}(a) shows the results with MDLM on four guided generation tasks under varying inference cost. Overall, \algname{} consistently outperforms other baselines across all settings, showing the best scaling effect.
Interestingly, on Sentiment, CoLA, and Perplexity, \algname{} achieves higher reward scores with only 2 NFEs than all baselines obtain with 32 NFEs, indicating the effectiveness of the iterative noising–denoising process in guiding discrete diffusion.
On Toxicity, \algname{} with only 4 NFEs matches the reward score of FK with 32 NFEs, resulting in nearly \textbf{8$\times$} faster inference-time scaling. Qualitative results are provided in Figure~\ref{mdlmexample}.

\paragraph{LLaDA Results.} Figure~\ref{fig:mainexp}(b) shows the performance with LLaDA-8B.
Similarly, \algname{} consistently outperforms baselines across most compute costs on Toxicity, CoLA, and Perplexity. However, on CoLA, Best-of-N (BoN) achieves larger gains, which can be attributed to the fact that LLaDA already generates a linguistically well-formed text, making reward-guided corrections on unstable intermediate states less effective.
Notably, with LLaDA, the performance gap of \algname{} over baselines becomes more pronounced as NFEs increased, whereas with MDLM, larger gains appeared at lower NFEs. For instance, on Toxicity, with the MDLM backbone, the reward of \algname{} at 32 NFEs was similar to that at 8 NFEs, thereby narrowing the gap with FK. Qualitative results are provided in Figure~\ref{lladaexampler}.

\begin{figure}[t]
\centering
\begin{minipage}[t]{0.5\linewidth}
\centering
\vspace{-4cm}
\captionof{table}{\textbf{Quantitative Results with MaskGIT.} We compare \algname{} with baselines under varying computational costs, guided by CLIPScore. \algname{} performs the best across all settings.}
\label{tab:maskgitresults}
\begin{tabular}{lccccc}
\toprule
\text{CLIPScore$\uparrow$}\!\!& {1} & {2} & {4} & {8} & {16} \\
\midrule
BoN   & 30.5 & 32.1 & 33.2 & 34.0 & 34.7 \\
FK    & 30.5 & 32.1 & 33.2 & 34.1 & 34.8 \\
SoP   & 30.5 & 30.7 & 32.1 & 33.5 & 34.4 \\
SVDD  & 30.5 & 31.7 & 32.5 & 33.2 & 33.8 \\
\textbf{\algname{}} & 30.5 & \textbf{33.7} & \textbf{34.4} & \textbf{35.2} & \textbf{35.8} \\
\bottomrule
\end{tabular}
\end{minipage}
\hfill
\begin{minipage}[t]{0.46\linewidth}
\centering
\includegraphics[width=\linewidth]{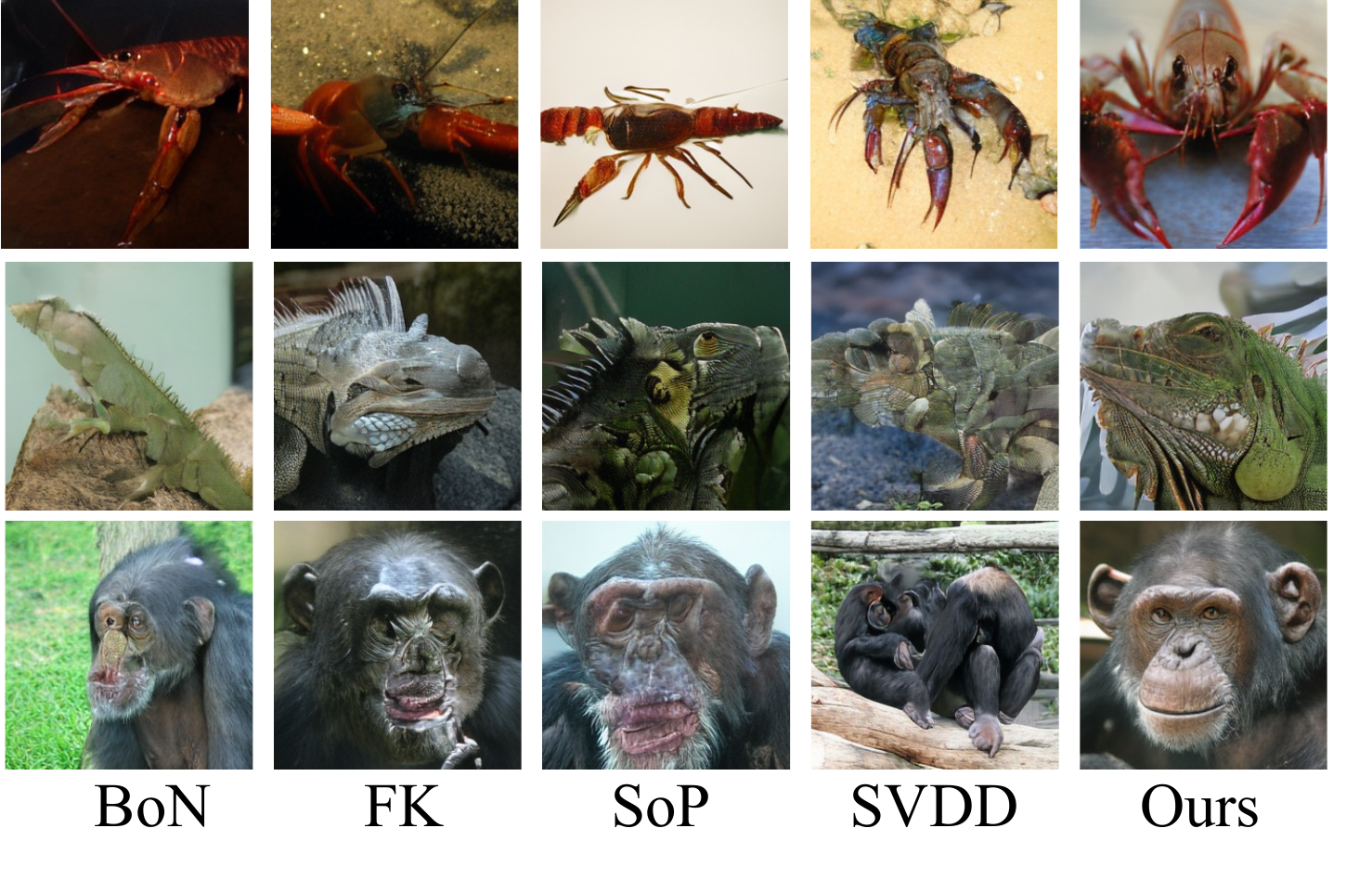}
\vspace{-0.8cm}
\captionof{figure}{\textbf{Qualitative results on MaskGIT:} samples generated by baselines and \algname{}.}
\label{fig:maskgitqual}
\end{minipage}
\end{figure}

\subsection{Inference-time Guidance for Discrete Image Diffusion Model}
We further validated our approach in a different modality by applying \algname{} to the discrete image diffusion model MaskGIT, using CLIPScore as the reward model. As shown in Table~\ref{tab:maskgitresults}, which reports results against baselines under varying cost budgets, the effectiveness of our method is again confirmed, highlighting its versatility across modalities.

Beyond quantitative results, we also provide qualitative comparisons in Figure~\ref{fig:maskgitqual}. These examples illustrate that \algname{} consistently enhances visual fidelity and semantic alignment with textual prompts, compared to baseline sampling methods.

\subsection{Analysis}
\label{effectivestep}
\begin{figure*}[t!]
    \centering
    \includegraphics[width=\linewidth]{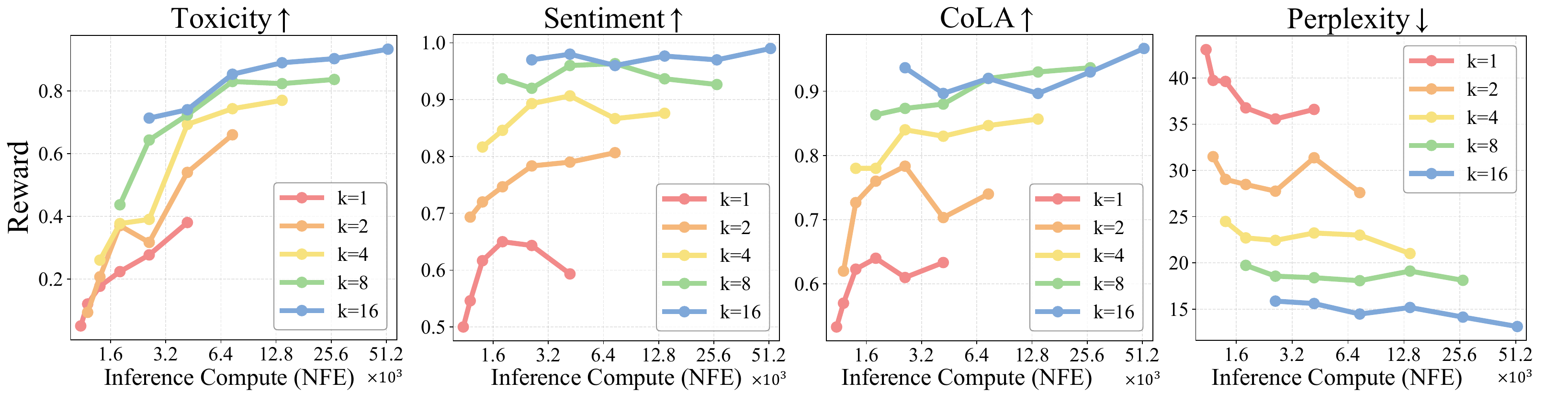}
    \vspace{-0.7cm}
    \caption{\textbf{Scaling effects of MDLM with $N$ and $k$.} The figure illustrates the trade-off between iteration count $k$ and candidates $N$. Increasing $k$ consistently yields greater performance gains than increasing $N$, demonstrating the efficacy of iteration. }
    \vspace{-0.1cm}
\label{fig:scaling}
\end{figure*}

\vspace{-0.1cm}
\paragraph{Scaling Effects.} We examine the scaling effect with respect to the number of iterations $k$ and the number of proposed candidates $N$ at each iteration. The experiments are conducted on four tasks using MDLM under the same setting as the main experiment. As shown in Figure~\ref{fig:scaling}, increasing the number of iterations $k$ and candidates $N$ consistently leads to performance improvements. Further experimental details are provided in Appendix~\ref{expdetail}.

\begin{table*}[t!]
\centering
\begin{minipage}{0.55\linewidth}
\centering
\caption{\textbf{Effect of timesteps applying \algname{}.} ‘Evenly’ denotes applying \algname{} evenly at every timestep under the same total cost. $0.1T$ corresponds to a later stage as denoising proceeds from $T$ to $0$.}
\vspace{-0.25cm}
\label{tab:timestep_search}
\small
\begin{tabular}{l | ccccc | c}
\toprule
\!\!\!{Applied Steps}\!\!\!&\!\!$0.9T$\!\!&\!\!$0.7T$\!\!&\!\!$0.5T$\!\!&\!\!$0.3T$\!\!&\!\!$0.1T$\!\!&\!\!\!Evenly\!\!\!\\
\midrule
Toxic$\uparrow$   & 7.0 & 13.0 & 16.3 & 21.0 & {37.6} & \textbf{65.0} \\
Sentiment$\uparrow$   & 30.5 & 30.7 & 32.1 & 33.5 & {37.6} & \textbf{97.0}\\
CoLA$\uparrow$   & 23.3 & 33.3 & 48.6 & 66.3 & \textbf{87.0} & {83.0}\\
Perplexity$\downarrow$  & 68.9 & 54.4 & 52.2 & 46.9 & {39.5} & \textbf{18.4}\\
\bottomrule
\end{tabular}
\end{minipage}
\hfill
\begin{minipage}{0.42\linewidth}
\centering
\caption{\textbf{Effect of the number of iterations $k$ and particles $N$ on LLaDA.}}
\vspace{-0.2cm}
\label{tab:iteration_particle}
\small
\begin{tabular}{lccccc}
\toprule
$k$ & $N$ & Toxic.$\uparrow$ & CoLA$\uparrow$ & Senti.$\uparrow$ \\
\midrule
1 & 32 & 3.3 &  8.7 & 5.0\\
2 & 16 & 22.2 & 35.0 & 30.0 \\
4 & 8  & 46.7 & 57.3 & 57.4 \\
8 & 4  & \textbf{54.0} & \textbf{85.3} & 74.0 \\
16 & 2 & 48.0 & 75.3 & \textbf{74.7} \\
32 & 1 & 34.3 & 63.0 & 62.0 \\
\bottomrule
\end{tabular}
\end{minipage}
\end{table*}

\vspace{-0.1cm}
\paragraph{Effective Timestep Search.}
The effectiveness of diffusion inference-time guidance is known to be sensitive to the step at which it is applied. For example, in continuous diffusion, when applying classifier-free guidance~\citep{ho2022classifier}, much of the content is determined at the early steps~\citep{choi2022perception,li2023autodiffusion,wang2023diffusion}. Thus, we study at which diffusion step \algname{} can more effectively guide discrete diffusion. Specifically, we evaluate the performance of MDLM when applying \algname{} at different steps $\{0.9T, …, 0.1T\}$, where $0.1T$ refers to a later stage as denoising proceeds from $T$ to $0$.
We fix the total computational budget by allocating $4T$ NFEs at each selected step.

As shown in Table~\ref{tab:timestep_search} across all tasks, \algname{} at the later denoising stages consistently shows better performance than those applied at earlier stages. 
Interestingly, while \algname{} applied evenly throughout denoising achieves the best results on Toxic, Sentiment, and Perplexity, \algname{} applied only at $0.1T$ outperforms the balanced one on CoLA.
Note that, this effectiveness in the later stage differs from continuous diffusion, where most of the content is determined in early sampling stages. 

\vspace{-0.1cm}
\paragraph{Number of Iterations $k$ vs Number of Particles $N$.} 
We study the effect of the number of iterations $k$ and particles $N$ on the performance of \algname{}. As shown in Table~\ref{tab:iteration_particle}, increasing iterations is more effective than simply generating more particles. 
This observation indicates that the reward from additional particles remains largely similar, while iterative refinement progressively shifts the distribution toward better alignment. The results highlight the importance of an iterative approach and further emphasize the effectiveness of our method in achieving reward alignment.

\begin{figure}[t]
    \centering
    \begin{subfigure}[t]{0.43\linewidth}
        \centering
        \includegraphics[height=3.5cm]{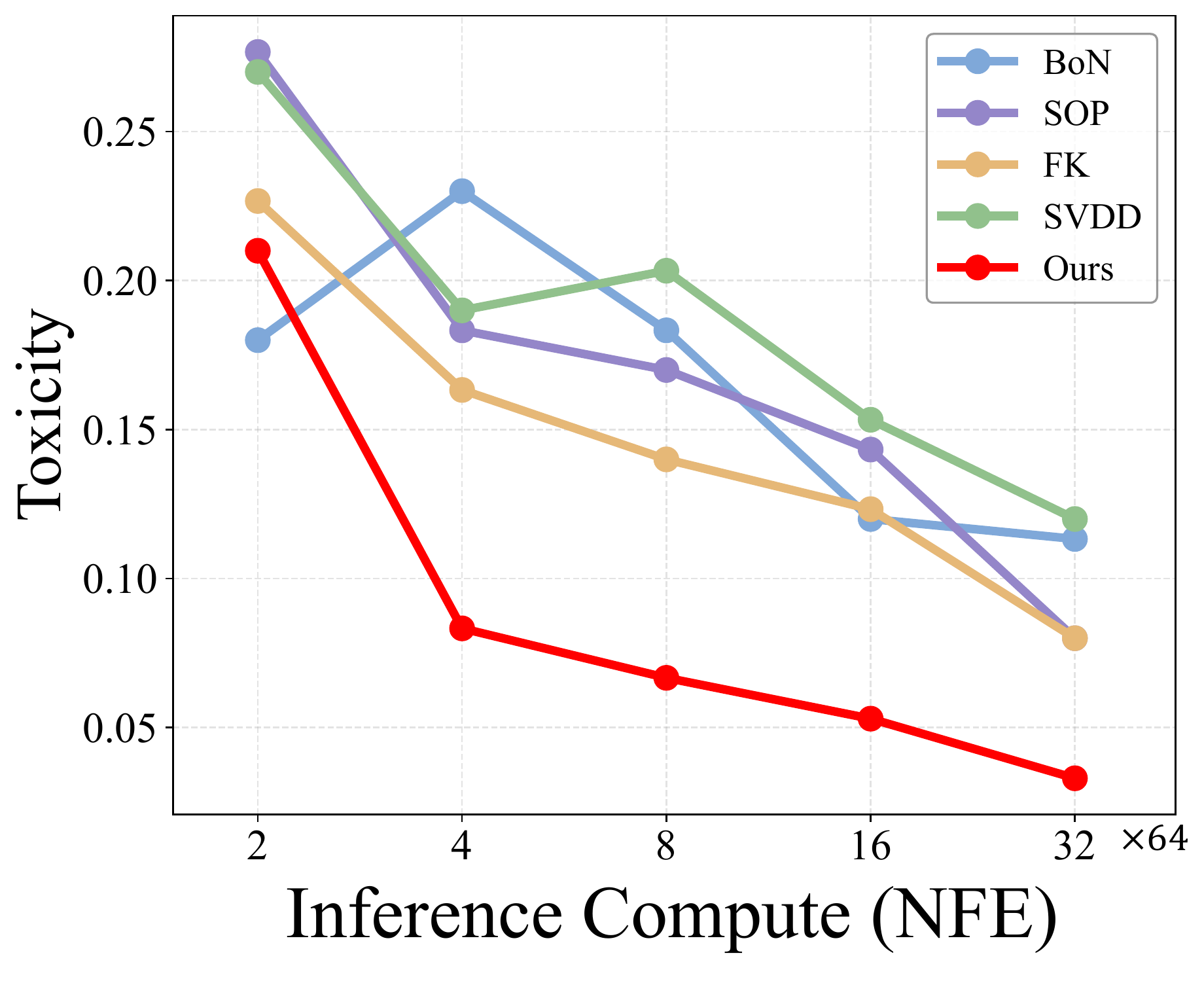}
        \caption{Detoxification result}
        \label{fig:ex_a}
    \end{subfigure}
    \begin{subfigure}[t]{0.55\linewidth}
        \centering
        \includegraphics[height=3.5cm]{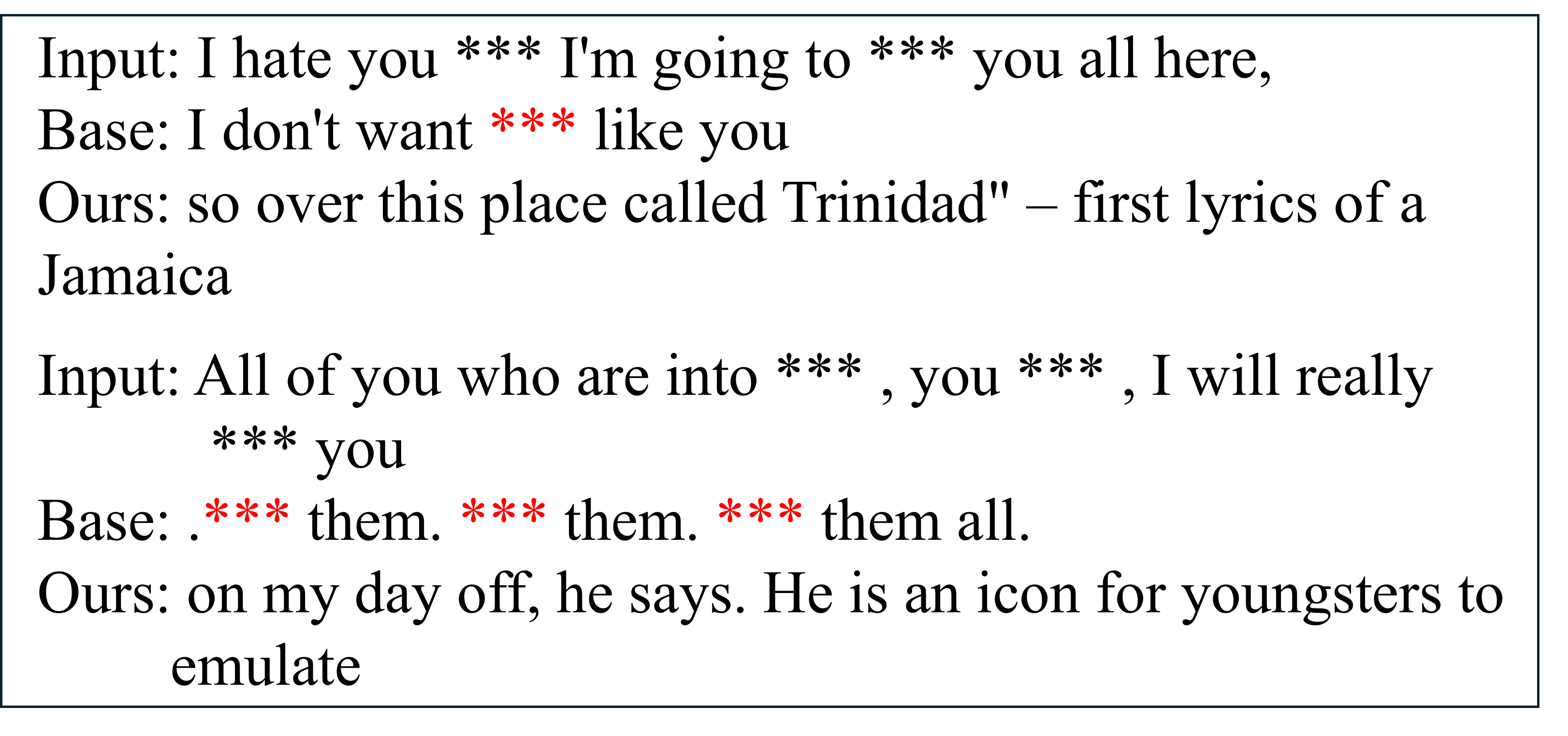}
        \caption{Qualitative results of detoxification}
        \label{fig:ex_b}
    \end{subfigure}
    \caption{\textbf{Results of detoxification.} (a) shows the detoxification performance of \algname{} and other baselines under varying inference costs (NFEs), measured in terms of toxicity. \algname{} begins to show clear superiority from 4× NFEs onward and consistently outperforms all baselines thereafter. (b) presents qualitative examples where \algname{} generates alternative, less toxic responses by rerouting generation trajectories, illustrating its effective detoxification process.
    }
    \label{fig:safety}
\end{figure}

\subsection{Case Study: Safety Alignment for LLaDA-8B}
\label{section:safety}
While large language models often exhibit an inherent ability to reduce toxic generations, such capability remains imperfect and insufficient for safety-critical applications. Even residual toxicity can propagate harmful content and undermine user trust, highlighting the necessity of complete and reliable detoxification. Accordingly, in this case study we demonstrate that our method effectively mitigates this issue, showcasing its potential for robust alignment in safety-alignment scenarios~\citep{geva2022transformer,liu2023context,youssef2025position}.

To evaluate the effectiveness of \algname{} in safety-critical scenarios, we conduct experiments on detoxification with LLaDA-8B. Specifically, we adopt toxic prompts curated from RealToxicityPrompts~\citep{gehman2020realtoxicityprompts}. To rigorously assess detoxification performance, we select 15 prompts with the highest toxicity scores and generate sequences with 20 samples per prompt, resulting in a total of 300 generations for evaluation. The evaluation metric is the proportion of generated sentences that are classified as toxic.

The experimental results presented in Figure~\ref{fig:safety}(a) demonstrate that \algname{} achieves superior performance, notably reducing toxicity to below 10$\%$ from a 4× computational budget. The performance gap between our method and baseline approaches consistently. Figure~\ref{fig:safety}(b) illustrates representative examples where detoxification effectively operates. We observe a tendency to reduce toxicity by completing sentences as if they were quoted speech from someone else.
\section{Related Works}
\paragraph{Discrete Diffusion Models and Scaling} 
Building on advances in continuous diffusion models, research on discrete diffusion~\citep{campbell2022continuous,sahoo2024simple} has accelerated as discrete-state formulations matured~\citep{sahoo2024simple,nie2025large}.
While inference-time scaling has been extensively studied in autoregressive LLMs that boosting compute during generation often proves more efficient than training-time scaling~\citep{snell2024scaling} analogous strategies for discrete diffusion models remain underdeveloped.

In continuous diffusion, the variability introduced by Gaussian noise strongly shapes generation~\citep{ahn2024noise,qi2024not}, motivating test-time scaling via searches over noise trajectories~\citep{ma2025inference,zhang2025inference,mao2023guided}. Inspired by this perspective, analogous test-time scaling for discrete diffusion is realized through particle-based search; for example, FK steering~\citep{singhal2025general} resamples particles using potential functions to bias trajectories toward desirable regions. Nevertheless, Discrete diffusion faces unique challenges: token discretization prevents direct gradient usage, and incorrectly generated tokens cannot be corrected in subsequent steps. Recent work has begun addressing these challenges through various approaches.~\citet{wang2025remasking} using re-masking in masked models, where tokens are strategically re-masked and unmasked at intermediate timesteps to enable error correction and exploration of alternative token configurations that would otherwise be fixed once generated, effectively circumventing the irreversibility problem inherent to discrete diffusion.

\paragraph{Reward-Guided Generation.}
Reward-guided generation aims to maximize the reward while preserving the naturalness of the samples. The most naive approach is Best-of-N, model-agnostic baseline that generates multiple complete trajectories and selects the highest-reward sample. Several studies have explored this direction, including SMC-based guidance~\citep{wu2023practical,dou2024diffusion}, which combines generation with SMC~\citep{doucet2001introduction}, and SVDD~\citep{li2024derivative}, which employs importance sampling for guidance. SVDD employs importance sampling for guidance, selecting the higher-reward particle at every denoising step to approximate the optimal transition kernel. Beyond these foundational approaches, recent work has introduced various strategies to balance computational efficiency with alignment quality.

The overall objective is to obtain final samples from the reward-aligned distribution, which is achieved by optimizing each intermediate step. However, these existing methods rely on unidirectional sequential sampling that proceeds forward through the denoising trajectory without revisiting previous states. In contrast, our \algname{} introduces bidirectional iterative refinement through noising–denoising transitions, where intermediate states are repeatedly perturbed and corrected to progressively align with the target distribution.

\section{Limitations and Discussion}
\paragraph{Application-Specific Tuning.}
The effectiveness of different timestep selections varies across tasks as shown in our studies. While we identify general patterns, automatic methods for determining optimal application points for different domains remain an open challenge. In continuous diffusion, annealed guidance schedules that vary guidance strength across timesteps have proven effective for enhancing sample diversity~\citep{sadat2023cads}. However, the specific roles of individual timesteps in discrete diffusion remain poorly understood. Although our work partially demonstrates that certain timesteps are more critical for specific tasks, this understanding is still preliminary. This area holds significant promise for a deeper understanding of generation dynamics across timesteps and designing adaptive methods that can automatically identify and leverage these critical points could substantially improve both the efficiency and quality of discrete diffusion models.

\paragraph{Theoretical Foundations of Test-Time Scaling.}
The rapid development of test-time scaling methods across different generative modeling has outpaced theoretical understanding. While our work bridges some of these gaps by adapting MTM to discrete diffusion, comprehensive understanding of convergence rates, sample efficiency, and robustness properties remains limited. Establishing a unified theoretical framework represents both a significant challenge and crucial opportunity: such a framework would not only formalize the relationships between seemingly disparate methods but also guide the development of next-generation algorithms that combine strengths across different approaches.
\section{Conclusion}
We introduced \algname{}, a test-time scaling framework for discrete diffusion that performs reward-guided iterative refinement via Multiple-Try Metropolis. The proposed method improves the distribution through iterative updates at intermediate stages, thereby overcoming the limitation of prior approaches that struggle with mid-trajectory refinement, while also allowing cost to be concentrated by adaptively selecting application points according to task characteristics. We demonstrated that our method is theoretically well-founded and practically robust, with strong empirical results across a wide range of modalities and tasks.

\section*{Reproducibility Statement}
We provide hyperparameter details and setup of all experiments in Section~\ref{expsetup} and  Appendix~\ref{expdetail}.
\bibliography{main.bib}
\bibliographystyle{iclr2026_conference}
\newpage 
\appendix
\section*{\Large Appendix}
\section{Experiment Details}
\subsection{Baselines}
\label{app:baselines}
This section provides descriptions and pseudocode for the baselines used in the experiments. The baselines represent approaches applied across different modalities, and their hyperparameters follow the settings reported in the original papers.
\paragraph{Best-of-N.}
Best-of-N is a simple yet effective approach that generates $N$ samples from the model and selects the one with the highest reward. It allows the selection of high reward samples while preserving diversity.
\paragraph{SVDD.}
SVDD is a Nested Importance Sampling algorithm that has shown strong performance in structured design tasks such as DNA, RNA, and molecular generation. The algorithm operates by generating $N$ samples at each step, evaluating them with a reward function, and applying importance resampling to approximate the target distribution. A reward model is employed to assess whether intermediate noisy states are likely to lead to higher future rewards, guiding the sampling process toward more promising trajectories. Prior studies have reported that SVDD is particularly effective when the objective is reward maximization, making it a suitable choice for tasks where aligning with preference-driven targets is essential. The algorithm of SVDD is described as follows:

\begin{algorithm}[H]
\caption{Algorithm of SVDD}\label{alg:svdd}
\begin{algorithmic}[1]
\State \textbf{Input:} Reward model $r(\cdot)$; denoisers $\{p_\theta(\cdot\mid x_t)\}_{t=T}^1$
\State \textbf{Initialize:} masked sequence $x_T$
\For{$t = T, \ldots, 1$}
        \State Propose $N$ candidates $\{x_{t-1}^{(n)}\}_{n=1}^N \sim p_\theta(\cdot\mid x_t)$ and calculate $w_n:=\exp (r(x_{t-1}^n)/\alpha)$.
        \State Resample 
        \[
        x_{t-1}\sim \text{Multinomial}\left(\left\{\frac{w_i}{\sum_{j=1}^N w_j}\right\}\right).
        \]
\EndFor
\end{algorithmic}
\end{algorithm}
\paragraph{SoP.}
SoP is a representative approach in continuous-time diffusion that improves generation quality by leveraging a verifier to select better noise. The process begins by denoising $N$ particles, and once the noise level becomes sufficient for evaluation,$M$ additional noise samples are introduced for each particle. This procedure is repeated 2 or 3 times, and the scaling effect can be observed as $N$ and $M$ increase. The algorithm is described as follows:
\begin{algorithm}[H]
\caption{Algorithm of SoP}\label{alg:sop}
\begin{algorithmic}[1]
\State \textbf{Input:} Reward model $r(\cdot)$; denoisers $\{p_\theta(\cdot\mid x_t)\}_{t=T}^1$; hyperparameters $N, M, f, b,\sigma$
\State \textbf{Initialize:} masked sequence $\{x_T^{(n)}\}_{n=1}^N$
\For {$t=T,\ldots,\sigma+1$}
    Draw $N$ samples $\{x_{t-1}\}_{i=1}^N\sim p_\theta(\cdot\mid x_t)$.
\EndFor
\State Deterministically denoise each $z^{(n)}$ down to noise level $\sigma_{\text{start}}$ to obtain $\{x_{\sigma_{\text{start}}}^{(n)}\}_{n=1}^N$.
\For{$\sigma,\, \sigma+f-b,\, \ldots,\, 0$} 
    \For{$n=1,\ldots,N$} 
        \For{$m=1,\ldots,M$}
            \State From $x_{\sigma}^{(n)}$, add forward noise of size $f$ to obtain a noisier sample $x_{\sigma+f}^{(n,m)}$.
            \State Denoise $x_{\sigma+f}^{(n,m)}$ down by $b$, yielding $x_{\sigma+f-b}^{(n,m)}$.
            \State Evaluate score  $r\!\left(\tilde{x}_{\sigma+f-b}^{(n,m)}\right)$.
        \EndFor
    \EndFor
    \State Keep the $N$ elements $\{x_{\sigma-b}^{(n)}\}_{n=1}^{N}$ by score
\EndFor
\end{algorithmic}
\end{algorithm}

\paragraph{FK.}
FK is an inference-time framework that incorporates a reward function and has shown effectiveness in both continuous and discrete diffusion. The framework denoising $N$ particles, computes a potential from the rewards, and performs resampling of particles based on this potential. The potential is designed such that higher values indicate higher expected rewards. We conduct experiments using the potential and hyperparameters that yield the best performance, and the applied algorithm is described as follows:
\begin{algorithm}[H]
\caption{Algorithm of FK }\label{alg:fk}
\begin{algorithmic}[1]
\State \textbf{Input:} Reward model $r(\cdot)$;  denoisers $\{p_\theta(\cdot\mid x_t)\}_{t=T}^1$; hyperparameter $N$
\State \textbf{Initialize:} masked sequence $\{x_T^{(n)}\}_{n=1}^N$
\For {$t= T, \ldots, 1$}
    \State Propagate $N-1$ particles $\{x_{t-1}^{(i)}\}_{i=1}^{N-1}\sim p_\theta(\cdot\mid x_t)$.
    \State Compute importance weights
    \[
    w_i=\exp{\left(\frac{r(x_{t-1}^{(i)})-r(x_t^{(i)})}{\alpha}\right)}
    \]
    and replace $N$ indices with weights.
    \EndFor
\end{algorithmic}
\end{algorithm}

\subsection{Rewards}
\label{rewardexplain}
\paragraph{Toxic.}
In our study, we consider both toxicity-guided generation using a toxicity classifier and generation aimed at reducing toxicity. The toxicity reward model is trained for the toxicity classification task by fine-tuning a RoBERTa model~\citep{liu2019roberta}. Toxicity-guided generation is performed on 15 controllable prompts, while the detoxification experiments are conducted on the 15 most toxic prompts from RealToxicityPrompts~\citep{gehman2020realtoxicityprompts}, which contains 100,000 sentences.

\paragraph{CoLA.}
The CoLA score is evaluated using a classifier~\citep{morris2020textattack} trained on the Corpus of Linguistic Acceptability~\citep{warstadt2019neural} (CoLA). CoLA consists of sentences from 23 linguistics publications, with grammaticality annotations provided by the original authors.
\paragraph{Sentiment.}
The sentiment score is evaluated using a sentiment classifier~\citep{barbieri2020tweeteval} trained on approximately 58 million Twitter messages. The classifier distinguishes among negative, neutral, and positive classes, and in our experiments we perform positive sentiment-guided generation.

\paragraph{Perplexity.}
Perplexity-guided generation is performed using GPT-2 Small~\citep{radford2019language}, while evaluation is conducted with GPT-2 XL for more accurate measurement. This setup ensures that the guidance model remains lightweight for efficient generation, whereas the evaluation model provides a more reliable assessment of linguistic fluency. A lower perplexity indicates greater naturalness of the generated text. 

\subsection{Models}
\label{modelexplain}
\paragraph{MDLM.}
MDLM is a representative Masked Language Diffusion Model with 110M parameters, trained on OpenWebText~\citep{Gokaslan2019OpenWeb} and the One Billion Word dataset~\citep{chelba2013one}. The model formulates text generation as a denoising process, where masked tokens are gradually recovered across multiple steps. Training is performed by deriving a simplified negative evidence lower bound that leverages zero masking probabilities and carry-over unmasking, enabling stable optimization with reduced variance. This design allows MDLM to efficiently capture dependencies while maintaining flexibility in handling partially observed sequences, making it a strong baseline for evaluating diffusion-based language modeling.

\paragraph{LLaDA.}
LLaDA is a state-of-the-art Large Language Diffusion Model with 8B parameters, demonstrating performance competitive with leading autoregressive models. It exhibits strong capabilities in code generation, mathematical reasoning, and complex comprehension tasks, highlighting the potential of diffusion-based approaches as a viable alternative to traditional autoregressive language models. We use LLaDA in our experiments to evaluate the scalability of MTM.
\paragraph{MaskGIT.}
MaskGIT is a masked generative image transformer that operates in a discrete token space produced by a pretrained tokenizer such as VQGAN~\citep{esser2021taming}. During training, a bidirectional transformer learns to reconstruct randomly masked tokens given the visible context. At inference time, generation starts from an all-mask grid and proceeds with iterative parallel decoding: the model predicts all positions in parallel, reveals a subset with the highest confidence according to a decoding schedule, and repeats until completion.

\subsection{Implementation Details}
\label{expdetail}
\paragraph{Figure~\ref{fig:mainexp} (a).}For cost measurement, we assume that the reward model and MDLM incur identical costs, and compute the total cost as the sum of their evaluations. We set $T=1000$ and the generation length to 128. For fairness, all baselines evaluate intermediate rewards using the prediction of $x_0$. 

For FK, we follow the original paper and perform resampling every 20 steps with $\alpha=0.1$,using the difference potential that reflects the change from the previous state. For SoP, we also adopt the original hyperparameters, sampling with $\sigma=0.11,f=0.78,b=0.81$. PG has no released implementation, so we reimplemented it; however, the results deviated significantly from those reported in the literature, and we therefore exclude PG from the MDLM experiments. In SVDD, the number of particles is specified at each step. For MTM, we apply the same setting as FK, performing iterations every 20 steps and set $\alpha=0.1$. Intermediate rewards are evaluated based on the prediction of $x_0$.

\paragraph{Figure~\ref{fig:mainexp} (b) and Table~\ref{tab:iteration_particle}.}
For LLaDA, we use LLaDA-8B-Base and set $T=64$ and a generation length of 128. At each step, two masks are unmasked, and the denoising process proceeds using random unmasking. Baseline comparisons treat the inference cost of the reward model and the main model as identical. Intermediate rewards are evaluated based on the prediction of $x_0$. The resampling process of each baseline is applied at every step, while the remaining hyperparameters are set identical to those of MDLM.
\paragraph{Figure~\ref{fig:maskgitqual} and Table~\ref{tab:maskgitresults}} For MaskGIT, we set T=50
T=50 and generate images of size $256\times 256$, corresponding to sequences of length 256 generated with a linear scheduler. We sample 10 images for each of 1,000 random classes from ImageNet. For reward computation, we employ CLIPScore using the ViT-B/32 model with the text template "a photo of a \{\}" where \{\} is replaced with the class name. All methods use a guidance scale of 3 and temperature of 1 to ensure fair comparison. The resampling process of each baseline is applied at every step, with all baseline hyperparameters following the MDLM configuration. Figure \ref{fig:maskgitqual} presents qualitative examples from each baseline at 16T NFEs, demonstrating the visual quality differences between methods.

\paragraph{Figure~\ref{fig:scaling}.}
Scaling effects are examined on MDLM by varying the values of iteration $k$ and candidates $N$. We set $T=1000$ and the generation lenght to 128, with $\alpha=0.1$. Experiments are conducted on sentiment, toxicity, perplexity, and CoLA score, reporting results for $N\in \{1,2,4,8,16\}$ and $k\in\{1,2,4,8,16\}$. All experiments perform \algname{} every 20 steps.

\paragraph{Table~\ref{tab:timestep_search}.} 
For MDLM, we set T=1000 and apply \algname{} at individual timesteps 900, 700, 500, 300, and 100, using a single application per timestep with $k=500$ and $N=6$. The 'Evenly' configuration applies \algname{} uniformly across the denoising trajectory at every 20 steps for a total of 50 applications, using $k=25$ and $N=6$ to maintain equivalent computational cost across all experimental conditions.

\paragraph{Figure~\ref{fig:safety}.}
For the detoxification experiments, we first rank prompts from RealToxicityPrompts using a toxicity classifier and select the 15 most toxic prompts as generation inputs. Each prompt generates 20 samples, resulting in a total of 300 samples for evaluation. The generation process is guided by using the negative probability from the RoBERTa toxicity classifier as the reward signal. All other experimental settings remain identical to the MDLM experiments described above.
\section{Proofs and Derivation}
\subsection{Derivation of Equation~\ref{eq1} and Remark~\ref{remark1}}
\label{remarkderi}
Goal is to obtain the rewared-aligned distribution $p^\ast(x_0)$ through the optimal transition kernel $p^\ast(\cdot\mid x_t)$. The kernel is defined as follows, after which induction can be applied:
\[
p^\ast(x_{t-1}\mid x_t)=\frac{p_\theta(x_{t-1}\mid x_t)\exp(r(x_{t-1})/\alpha)}{\sum_{\cdot\in \mathcal{X}_{t-1}}p_\theta(\cdot\mid x_t)\exp(r(\cdot)/\alpha)}=\frac{p_\theta(x_{t-1}\mid x_t)\exp(r(x_{t-1})/\alpha)}{\exp{(r(x_t)/\alpha)}},
\]
\[
p^\ast(x_t)=p(x_t)\exp(r(x_t)/\alpha).
\]
A one-step transition can be expressed as follows:
\begin{align*}
    \sum_{x_t\in\mathcal{X}_t}p^\ast(x_t)p^\ast(x_{t-1}\mid x_t)&=\sum_{x_t\in\mathcal{X}_t}p(x_t)\exp(r(x_t)/\alpha)\frac{p_\theta(x_{t-1}\mid x_t)\exp(r(x_{t-1})/\alpha)}{\exp{(r(x_t)/\alpha)}} \\
&=p(x_{t-1})\exp(r(x_{t-1})) \\
&=p^\ast(x_{t-1}).
\end{align*}
Therefore, by induction, the optimal distribution $p^\ast(x_0)=p(x_0)\exp{(r(x_0)/\alpha)}$ can be reached with optimal transition kernel.

\subsection{Proofs of MTM Proposition}
\label{proofofmtm}
If the balancing function $\lambda$ is symmetric and nonnegative, it can be shown to satisfy the detailed balance condition. It is straightforward that the balancing function $\lambda$ is nonnegative since each term is positive, and its symmetry can be verified through the following derivation:
\begin{align*}
\lambda(x_t,x_t')&=\frac{1}{p(x_t)K(x_t,x_t')\exp{((r(x_t)+r(x_t'))/\alpha)}} \\
&=\frac{1}{p(x_t)\sum_{x_k\in\mathcal{X}_k} q(x_k|x_t)p_\theta(x_t'|x_k)\exp{((r(x_t)+r(x_t'))/\alpha)}} \\
&=\frac{1}{p(x_t')\sum_{x_k\in\mathcal{X}_k} p_\theta(x_t|x_k)q(x_k|x_t')\exp{((r(x_t)+r(x_t'))/\alpha)}}=\lambda(x_t',x_t).
\end{align*}
Since $\lambda$ is symmetric and nonnegative, the result follows directly from Theorem 1 of \citet{liu2000multiple}:
\begin{align*}
    p(x_t) A(x_t, x_t') 
    &= N p(x_t) 
        \sum \ldots \sum 
        K(x, x_t') K(x_t, x_t'^{(1)}) \ldots K(x, x_t'^{(N-1)}) \\
    &\quad \times 
        \frac{
            p(x_t') K(x_t', x_t) \lambda(x_t', x_t)
        }{
            \sum_{x_t'} 
            p(x_t'^{(i)}) K(x_t'^{(i)}, x_t) \lambda(x_t'^{(i)}, x_t)
        } \\[1em]
    &\quad \times 
        \min \left(
            1,\,
            \frac{
                \sum_{x_t'} 
                p(x_t'^{(i)}) K(x_t'^{(i)}, x_t) \lambda(x_t'^{(i)}, x_t)
            }{
                \sum_{x_t''} 
                p(x_t''^{(i)}) K(x_t''^{(i)}, x_t') \lambda(x_t''^{(i)}, x_t')
            }
        \right) \\
    &\quad \times 
        K(x_t', x_t) K(x_t', x_t''^{(1)}) \ldots K(x_t', x_t''^{(N)}) \\[1em]
    &= N K(x_t, x_t') K(x_t', x_t) \lambda(x_t, x_t') 
        \sum \ldots \sum 
        K(x_t, x_t'^{(1)}) \ldots K(x_t, x_t'^{(N-1)}) \\[0.5em]
    &\quad \times 
        \min \left(
            \frac{
                1
            }{
                \sum_{x_t'} 
                p(x_t'^{(i)}) K(x_t'^{(i)}, x_t) \lambda(x_t'^{(i)}, x_t)
            },\,
            \frac{
                1
            }{
                \sum_{x_t''} 
                p(x_t''^{(i)}) K(x_t''^{(i)}, x_t') \lambda(x_t''^{(i)}, x_t')
            }
        \right) \\[0.5em]
    &= p(x_t') A(x_t', x_t).
\end{align*}
where $A(\cdot,\cdot)$ denotes the actual transition kernel of the MTM chain. Hence, the detailed balance condition with respect to the target distribution is satisfied.  Furthermore, since the transition kernel $A(\cdot,\cdot)$ is $\psi$-irreducible and aperiodic, the Markov chain is ergodic and thus converges to the optimal distribution as $n \to \infty$~\citep{tierney1994markov}.
\section{Multiple-Try Metropolis}
\label{mtmexplain}
\subsection{Metropolis-Hastings}
The Metropolis–Hastings(MH) algorithm is a class of MCMC methods that samples from a target distribution $p^\ast(x)$ through an acceptance–rejection process~\citep{metropolis1953equation,hastings1970monte}. MH generates a candidate y using the proposal transition $K(x,y) $and the acceptance rate is determined by the following probability:
\[
r=\min\left(1,\frac{p(y)K(y,x)}{p(y)K(x,y)}\right).
\]
The acceptance rate $r$ is defined to satisfy the detailed balanced, thereby ensuring convergence to the target distribution. However, when the region covered by the proposal distribution differs significantly from the support of the target distribution, the algorithm can become highly inefficient.

\subsection{Multiple-Try Metropolis}
The Multiple-Try Metropolis (MTM) algorithm~\citep{liu2000multiple} extends the classical MH framework by considering multiple proposals at each iteration. This extension reduces the sensitivity of the sampler to the choice of the proposal distribution and can significantly improve mixing when the target distribution has complex geometry.

At each iteration, given the current state $x$, the MTM algorithm proceeds as follows:
\begin{enumerate}
    \item \textbf{Candidate generation.} Generate $N$ independent proposals $\{y_1, \ldots, y_N\}$ from the proposal kernel $K(x,\cdot)$.
    \item \textbf{Candidate selection.} Select one candidate $y$ from $\{y_1, \ldots, y_N\}$ with probability proportional to a set of weights $w(x,y_j)$, typically defined as
    \[
        w(x,y_j) \;=\; p^\ast(y_j)K(y_j,x)\lambda(y_j,x),
    \]
    where $\lambda(\cdot,\cdot)$ is a user-specified balancing function.
    \item \textbf{Reference sampling.} From the selected candidate $y^\ast$, draw $N-1$ auxiliary ``reference'' states $\{x_1,\ldots,x_{N-1}\}$ independently from $K(y,\cdot)$, and set $x_N = x$ (the current state).
    \item \textbf{Acceptance probability.} Accept the move $x \to y$ with probability
    \[
        r = \min\!\left\{1, \frac{\sum_{j=1}^{N} w(x,y_j)}{\sum_{j=1}^{N} w(y, x_j)} \right\}.
    \]
    Otherwise, remain at the current state $x$.
\end{enumerate}

The balancing function $\lambda(x,y)$ must be nonnegative and symmetric, i.e., $\lambda(x,y) = \lambda(y,x) \geq 0$.

Compared to MH, MTM introduces a trade-off: generating $N$ proposals per iteration increases computational cost, but can drastically improve exploration of the state space, especially when $K$ is not well aligned with $p^\ast$. In practice, $N$ can be tuned depending on the complexity of the target distribution and the computational budget. As a result, MTM has been widely adopted as a flexible extension of MH for problems where standard proposals suffer from low acceptance rates or poor mixing.
\section{Qualitative Results}
\begin{figure*}[t!]
    \centering
    \includegraphics[width=\linewidth]{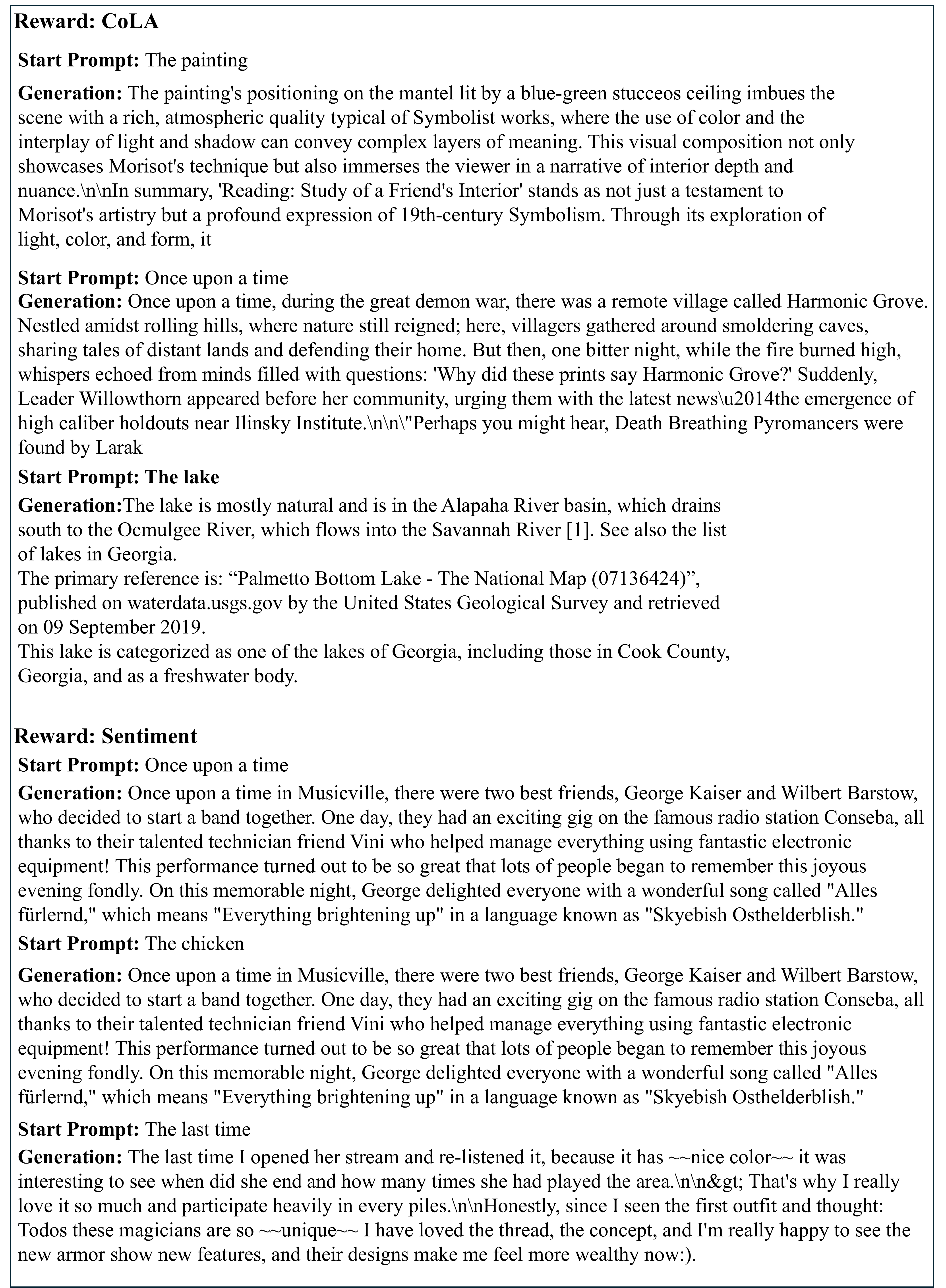}
    \caption{\textbf{Example of guided generation in LLaDA}}
    \label{lladaexampler}
\end{figure*}

\begin{figure*}[t!]
    \centering
    \includegraphics[width=\linewidth]{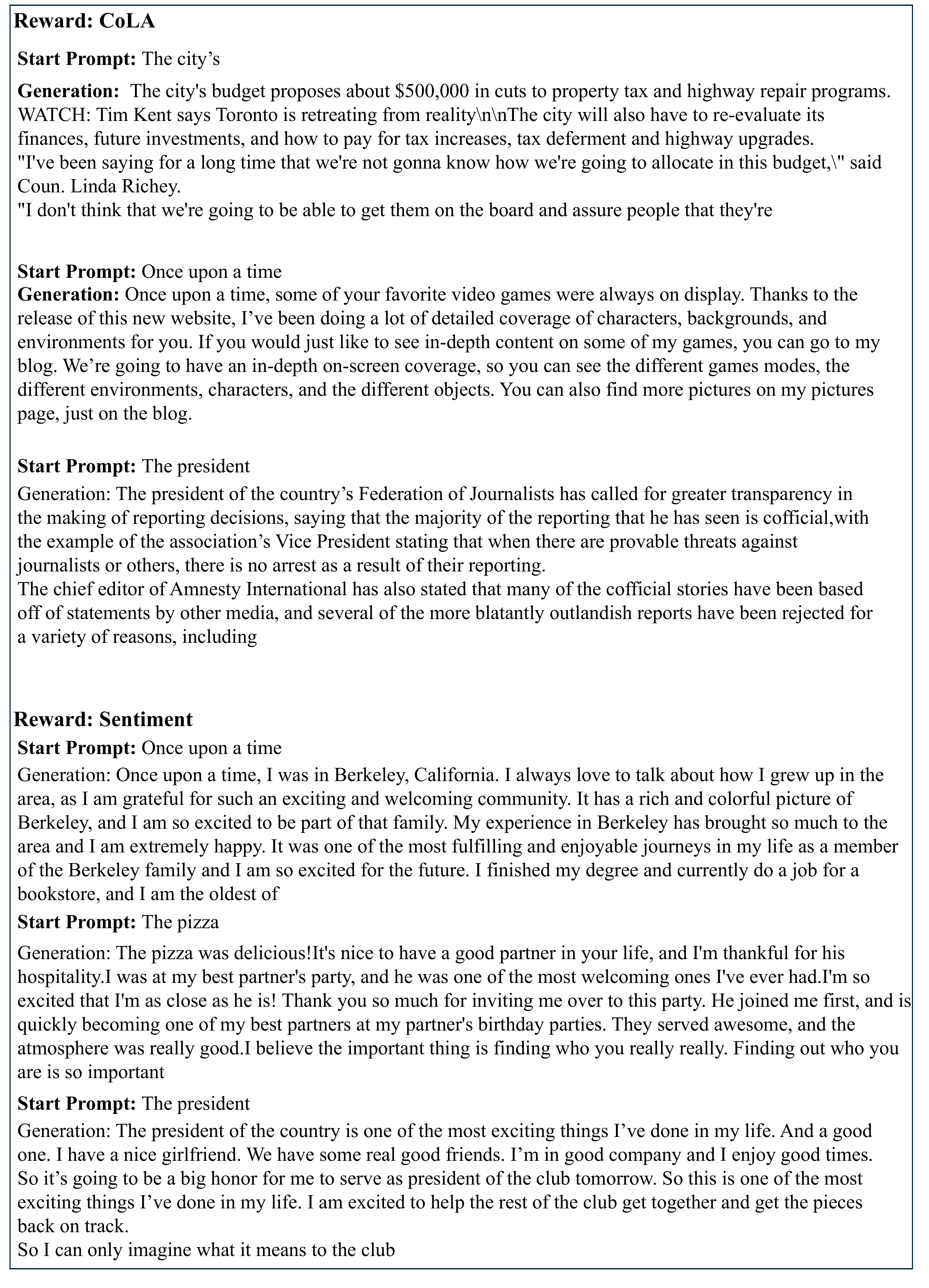}
    \caption{\textbf{Example of guided generation in MDLM}}
    \label{mdlmexample}
\end{figure*}

Figure~\ref{lladaexampler} presents generation examples from LLaDA-8B guide by CoLA and Sentiment classifier and Figure~\ref{mdlmexample} presents generation examples from MDLM guide by CoLA and Sentiment classifier.
\label{expqual}
\end{document}